\pdfoutput=1
\documentclass[11pt,xcolor={usenames,dvipsnames}]{article}

\usepackage[]{EMNLP2023}

\usepackage{times}
\usepackage{latexsym,amsmath,amsthm,amssymb,bm}
\RequirePackage[inline,shortlabels]{enumitem}
\setlist{nosep}
\RequirePackage{multirow}
\usepackage{booktabs}
\usepackage{adjustbox}
\usepackage{algorithm}
\usepackage{algpseudocode}
\usepackage{comment}
\usepackage[T1]{fontenc}
\usepackage[utf8]{inputenc}

\usepackage{microtype}

\usepackage{inconsolata}

\RequirePackage[textsize=scriptsize,color=yellow!10,linecolor=orange]{todonotes}
\makeatletter
\g@addto@macro{\normalsize}{%
\setlength{\abovedisplayskip}{2pt plus1pt}%
\setlength{\abovedisplayshortskip}{2pt plus1pt}%
\setlength{\belowdisplayskip}{2pt plus1pt}%
\setlength{\belowdisplayshortskip}{2pt plus1pt}}
\makeatother

\def\sysname{{\sffamily CoRE-CoG}}
\def\decodername{{\sffamily HopSkip}}

\def\ztitle{End to End Conversational Recommendation System}
\def\ztitle{\sysname: Conversational Recommendation of Entities using Constrained Generation}
\title{\ztitle}

\author{\begin{tabular}{c} Harshvardhan Srivastava$^{1}$, Kanav Pruthi$^{1}$, Soumen Chakrabarti$^{2}$, Mausam$^{1}$ \end{tabular} \\
\vspace{2mm}
$^1$IIT Delhi \hspace{2mm} $^2$IIT Bombay \\
\footnotesize\texttt{\begin{tabular}{c} \{srivastavahv, kanav.pruthi\}@gmail.com, soumen@cse.iitb.ac.in, mausam@cse.iitd.ac.in \end{tabular}}
}

\begin{document}
\maketitle

\begin{abstract}
End-to-end conversational recommendation systems (CRS) generate responses by leveraging both dialog history and a knowledge base (KB).  A CRS mainly  faces three key challenges:
\begin{enumerate*}[(1)]
\item at each turn, it must decide if recommending a KB entity is appropriate;
\item if so, it must identify the most relevant KB entity to recommend; and finally, 
\item it must recommend the entity in a fluent utterance that is consistent with the conversation history.
\end{enumerate*}
Recent CRSs do not pay sufficient attention to these desiderata, often generating unfluent responses or not recommending (relevant) entities at the right turn.
%Some recent CRSs generate utterances from language models without sufficient enforcement of a recommended entity mention; they tend to have lower recommendation performance.
%Others focus on entity selection but only indirectly encourage their language generation to include the entity mention in a fluent manner.
We introduce a new CRS we call \sysname{}. \sysname{} addresses the limitations in prior systems by implementing
\begin{enumerate*}[(1)]
\item a recommendation trigger that decides if the system utterance should include an entity,
\item a type pruning module that improves the relevance of recommended entities, and
\item a novel constrained response generator to make recommendations while maintaining fluency.
\end{enumerate*}
Together, these modules ensure simultaneous accurate recommendation decisions and fluent system utterances.
Experiments with recent benchmarks show the superiority particularly on conditional generation sub-tasks with up to 10 F1 and 4 Recall@1 percent points gain over baselines. 
\end{abstract}

\section{Introduction}
\label{sec:Intro}

Conversational recommendation systems (CRSs) incorporate natural language dialogue into the recommendation process \cite{crsSurvey}. Compared to traditional recommender systems, CRSs use conversation to probe deeper into user needs and preferences, clarify ambiguous queries, and elicit additional contextual information, all of which can enhance the relevance and personalization of recommendations. Recent benchmarks such as Durecdial~2.0 \citep{liu-etal-2021-durecdial} and ReDial \citep{redial} maintain a KG of entities and relationships, along with conversational data with users for training CRSs. Our goal is to build end-to-end CRSs, i.e., those systems which perform retrieval and response generation in the same system, without intermediate manual annotation.

%A CRS faces three key challenges. First, it needs to ascertain whether more user information is to be elicited or it can make a recommendation at the current turn.  For example, a system may respond with a general chitchat statement saying: “\textit{Many people like this hot dish!}” when discussing on the topic of food recommendation, but the more appropriate response would be "\textit{I recommend you to \textbf{Wanzhou Roasted Fish}  to eat \textbf{Marinated Fish}. You can book it.}" which explicitly mentions both the restaurant and the food recommendation. Second, if a recommendation is to be made, it needs to decide on the most relevant KB entity to recommend. E.g., during the conversation involvng closely related topics like movies and music, which may have shared entities, the system might mistakenly recommend a song by Jacky Cheung (say, \textit{\textbf{Wait Until Flowers Wither}}) instead of recommending his movie (say, \textit{\textbf{The Days of Being Dumb}}).  Finally, the recommendation should be made in a fluent and expressive manner that is consistent with conversation history. Continuing the running example, "\textit{By the way, Leslie Cheung's \textbf{Love Now} is also very good. It's very catchy. You can try it.}" is much expressive and informative than "\textit{\textbf{Love Now} is also good.}" which is produced by some constrained decoding strategy like Neurologic A*esque decoding \cite{neurologica*}.

A CRS faces three key challenges. Firstly, it needs to ascertain whether more user information is to be elicited or it can make a recommendation at the current turn. Secondly, if a recommendation is to be made, it needs to decide on the most relevant KB entity to recommend based on conversation context.  Finally, the recommendation should be made in a fluent and expressive manner that is consistent with conversation history. 
% \todo{M: I was not very happy with examples and we had less space. If Soumen feels otherwise he can bring them back in. Commented above.}

Recent CRSs \cite{mese,kgsf} do not give adequate attention to these challenges. Our work is built over a state-of-the-art system, MESE \cite{mese}, which combines three neural modules: entity retriever, entity reranker and a response generator.  Experiments with MESE suggest that it generates utterances from language models without strict enforcement of recommended entity mention --- consequently, it frequently misses recommending any entity at the appropriate turns.  Moreover, there is a sizeable number of mismatch errors in entity-type requested versus recommended.

In response, we present \sysname{} --- Conversational Recommendation of Entities
using Constrained Generation. \sysname{} makes three important modifications over the MESE architecture. First, it adds a recommendation trigger, which decides, based on the conversation history, the appropriateness of the CRS making a recommendation in its current turn. Second, it makes the retriever type-cognizant, improving the relevance of entity chosen for recommendation. Finally, and most importantly, it constrains the response generator (decoder) to include the recommended entity mention without fail (while encouraging fluency).  To the best of our knowledge, this is the first application of constrained generation in the dialog setting.

This last modification exposes a tradeoff between recommendation accuracy and fluency of conversational utterances. Existing constrained generation solutions \cite{lu2021neurologic,qin2022cold}, when applied in a CRS, generate less fluent responses. Since exactly one entity needs to be recommended in our case, we propose a novel bidirectional decoder (\decodername), where the utterance text is constructed \emph{around} the entity mention, such that it is both more fluent and consistent with the conversation history.

We compare \sysname{} against recent CRSs using ReDial and Durecdial~2.0 datasets.
We show that \sysname{} achieves high entity recommendation accuracy, while simultaneously generating recommendation utterances that are natural and fluent.
On Durecdial 2.0, \sysname{} with our novel \decodername{} outperforms other conditioned generation approaches by about 2.4 BLEU points, and existing unconditional generators by about 10 Entity F1 percent points. It also improves retriever performance by 4 Recall@1 percent points. We release the code and data for further research on the problem \href{https://drive.google.com/drive/folders/1LPHe8ugbnu3sLtpYbS-accMcVjhwGs1o?usp=drive_link}{here}.

\section{Related work}
\label{sec:Related}
A conversational recommender system (CRS), given a KB of facts and a dialog history, produces the next utterance, with the underlying goal of eliciting user needs and recommending the best entities from the KB. An end-to-end CRS is trained only on dialog data, without any additional intermediate supervision. A CRS is evaluated both on quality of recommendations (accuracy of retrieved entities), and that of utterances (fluency, naturalness, etc). 

Several end-to-end approaches exist for conversational recommendation.  KBRD \citep{kbrd} makes use of a self-attention transformer and relational GCN \citep{rgcn} to return a distribution over the entities that can be recommended at a given time. KGSF \citep{kgsf} incorporates a commonsense knowledge graph (ConceptNet) in addition to the entity-KG, to establish relations between words, such as the synonyms, and antonyms. It uses a mutual information maximization approach to bridge the semantic gap between words and entities. 
To handle incompleteness of KGs, CRFR~\citep{zhou-etal-2021-crfr} uses reinforcement learning over multiple incomplete paths to return best recommendations. CR Walker \citep{ma2021crwalker} uses multiple walks over the background KG starting from a single node to reach the target entity to be recommended.

Our work is based on MESE \citep{mese}, which we found to have the highest performance on existing datasets in our initial experiments. MESE learns three neural modules, all based on pre-trained language models (PLMs). An entity retriever encodes each KG entity using its metadata, and shortlists a few based on matching with dialog context. An entity reranker performs a collective reranking over the set of retrieved entities. The highest ranked entity is used as input in a response generator to output the system utterance. 

\label{sec:Relatedconst}
\textbf{Constrained generation approaches}: Constrained generation refers to generating a sequence output with an additional constraint, e.g., on length of output or presence/absence of a word. Since \sysname{} formulates response generation in this paradigm (utterance must have the entity token), we briefly survey existing constrained generation approaches. Neurologic A*esque  \citep{lu2021neurologic} defines a cost function that is associated with mentioning constraint words, and generates text while minimizing a look-ahead cost. COLD Decoding \cite{qin2022cold} performs decodiing by sampling from an energy-based model, plugging in constraint parameters, and then sampling the next word from the resultant distribution. PPLM \cite{pplm} uses a bag-of-words or one-layer classifiers as attribute controllers to exercise control over the attributes of generated text, while retaining fluency. All these are autoregressive models. Finally, Diffusion-LM \cite{li2022diffusionlm} is a non-autoregressive model that iteratively denoises a sequence of Gaussian vectors into word vectors, yielding latent variables used by a gradient-based algorithm for controlled generation.

\section{Notation and preliminaries}
\label{sec:Prelim}

Throughout, we use $a, u, e$ to denote words and entities, and $\bm{a}, \bm{u}, \bm{e} \in \mathbb{R}^D$ for their embedding vectors.  A sequence of symbols is denoted $\vec{a}$; the corresponding sequence of vectors is written as~$\vec{\bm{a}}$.  For an integer $N$, $[N]$ denotes $\{1,\ldots,N\}$.

\subsection{Knowledge base}

Apart from the conversation history, the CRS has access to a knowledge base (KB), denoted~$K$.
The KB consists of a set of entities $E$ and relations $R$, and some number of triples $(s,r,o)$ (subject, relation, object) where $s, o \in E$ and $r \in R$.
An important special relation is ``is-instance-of'', e.g., (True Lies, is-instance-of, movie).  Here `movie' is said to be the \emph{type} of the entity ``True Lies''.
Let $T$ denote the set of all types in the KB.
We will use this special relation for better entity selection.

\subsection{Entity embeddings}

Traditionally, KB (or KG, for Knowledge Graph) embeddings have been derived from the topology of the KG and its edge labels (relation IDs).  ComplEx \citep{Lacroix2018ComplExN3} and ConvE \citep{Chao+2019GcnConveKGC} are some of the effective KG embedding methods in use.  More recently, rich textual features associated with entities and relations have been harnessed in addition to graph structure \cite{Peters2019KnowBERT, zhang2019ernie, Yao2019kgbert}.
While these remain options in our setting, dialog data sets have associated KBs that are relatively small and simple in structure.  Therefore, we characterize each entity by its name(s), description and a selected group of attributes, concatenated with a suitable separator token. 
E.g., in the context of music recommendations, we gather information such as each music entity $e$'s title, singers, lyrics, and music genre.
This textual representation of each entity $e$ is then input to an text encoder, specifically, DistilBERT \citep{sanh2019distilbert}, which 
transforms the token sequence into a sequence of embedding vectors.  These vectors are then pooled into a single fixed-length vector, to which we apply one feed-forward layer to generate the output embedding $\bm{e}$ of the entity. We apply this process to precompute embeddings for all entities in the KB.

% Pooling is done using Key, Query and Value method -> Used in general transformers

\subsection{Problem definition}

We first describe the task of learning an end-to-end Knowledge Base (KB) assisted Conversational Recommendation System (CRS). Let
\begin{align}
H = (\langle \vec{u}_m, \vec{a}_m \rangle: m \in [M])
\end{align}
denote $M$-turn history of a dialog between a user and the system `agent'.
The $m$th user utterance
\begin{align}
\vec{u}_m = (u_{m,\ell}: \ell \in [U_m])
\end{align}
is a sequence of $U_m$ tokens.  Similarly, the $m$th agent utterance
\begin{align}
\vec{a}_m = (a_{m,\ell}: \ell \in [A_m])
\end{align}
is a sequence of $A_m$ tokens.
Given the conversation history $H$, the next user utterance $\vec{u}_{M+1}$, the goal is to produce the response utterance~$\vec{a}_{M+1}$.

\section{\sysname{} architecture}
\label{sec:Method}

In this section, we present the architecture of \sysname, which consists of the following components:
\begin{enumerate*}[(1)]
\item recommendation trigger,
\item entity type predictor,
\item entity reranker, and
\item a constrained decoder.
\end{enumerate*}
\sysname{} system first involves the recommendation trigger to identify the decision point for recommendations.  Next, the entity type predictor and reranker rank entities based on their types, given the conversation history.  
Although conceptually simple, the careful decision to recommend and type-sensitive selection of the entity to recommend substantially reduces error cases seen in baselines.
Finally, this selected entity is compulsorily included in an utterance $\vec{a}_{M+1}$ generated by a constrained decoder.

\subsection{Unified history representation $\vec{\bm{H}}$}
\label{notation:unifiedHistory}

\begin{figure}[t]
\centering\includegraphics[width=\hsize]{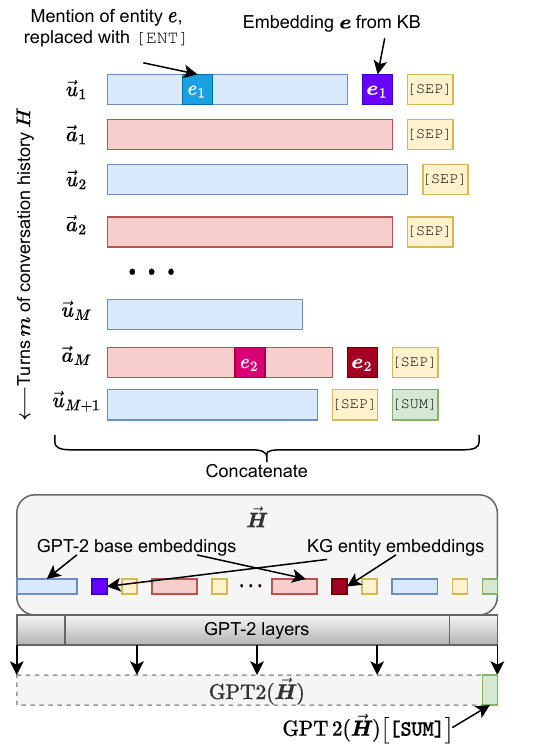}
\caption{Illustration of how $\vec{\bm{H}}$ is obtained from the conversation history and KB embeddings, and how GPT-2 is applied to it to form a shared representation of history, to be used by different heads downstream.}
\label{fig:UnifiedHistory}
\end{figure}

From $H$, $\vec{u}_{M+1}$, and the KB with entity embeddings, we build a unified representation $\vec{\bm{H}}$ of the conversation history, as follows.
First, all entity mentions in $H$ and $\vec{u}_{M+1}$ are replaced with a special token \texttt{[ENT]}. While entity resolution and disambiguation have sophisticated solutions \citep{laskar-etal-2022-blink}, for current dialog benchmarks, simple string matching suffices. If an entity mention span in $\vec{u}_m$ or $\vec{a}_m$ is replaced by \texttt{[ENT]}, we do remember the identity of the entity $e$ from the KB.
We use the base embedding matrix of a language model (LM) such as GPT-2 \cite{gpt2} to map each token $u_{m,\ell}$ or $a_{m,\ell}$ to the base embeddings $\bm{u}_{m,\ell}$ or $\bm{a}_{m,\ell} \in \mathbb{R}^D$.  At positions marked \texttt{[ENT]} we inject the base embedding of that special token.
This results in a sequence of base embeddings $\vec{\bm{u}}_m$ or $\vec{\bm{a}}_m$.
However, at the end of each such utterance sequence, we append the KB-based embeddings $\bm{e}$ of any entities $e$ that originally were mentioned in the utterance.
If the full history $H\|\vec{u}_{M+1}$ is too long (e.g., GPT-2 has a 1024-token input limit), we include only a suffix of the sequence (to retain the most recent turns).
At the very end of the sequence, we append the base embedding of another special token \texttt{[SUM]} (representing `summary') --- similar to \texttt{[CLS]} of BERT, but at the end.
We call this entire sequence of vectors $\vec{\bm{H}}$.
Figure\,\ref{fig:UnifiedHistory} illustrates with an example how $\vec{\bm{H}}$ is computed from a conversation history and the KB. For a backward generation pass as used in \ref{arch:hopskip}, we use a reversed form of history sequence $\operatorname{rev}(\vec{\bm{H}})$, where each utterance at previous timestep is reversed locally, but the relative ordering of the utterances is the same.

\subsection{Shared encoder}

We use GPT-2 (with all parameters to be fine-tuned) as a shared encoder for~$\vec{\bm{H}}$.
Its input is the unified conversation history~$\vec{\bm{H}}$, which ends with the base embedding of the special token \texttt{[SUM]} to represent a summary of the preceding sequence.  We pass this sequence of base embeddings through GPT-2 and read off the contextual output embedding at the \texttt{[SUM]} position:
\begin{align}
\operatorname{GPT2}(\vec{\bm{H}}) \big[ \texttt{[SUM]} \big] \in \mathbb{R}^D.
\label{eq:HistoryEmbed}
\end{align}
This will now be reused with different prediction heads in various modules.

\subsection{Recommendation trigger}

To comprehend the user's intent for system recommendations within a chatbot framework, developing a system behavior that can accurately identify the appropriate moment or timestep for providing recommendations becomes crucial. The need for system recommendations generally arises from two types of user behaviors. Firstly, when the user explicitly requests the system to respond with a recommendation, typically preceded by an explicit statement asking for a recommendation. However, during a natural conversation, it is often desirable for the system to autonomously identify the point at which a recommendation should be made based on the ongoing discussion from previous timesteps.

Our recommendation trigger is a simple sequence-to-0/1 classifier.
The output from the shared encoder is projected down to a scalar using a trained weight vector $\bm{w} \in \mathbb{R}^D$, and a sigmoid is applied --- this may be written as
\begin{align}
\sigma\Big(  \operatorname{GPT2}(\vec{\bm{H}}) \big[ \texttt{[SUM]} \big]  \cdot \bm{w} \Big)
\end{align}
From many training (`gold') conversations like $D$, for each prefix $M$, we mark where the agent actually output an entity recommendation vs.\ where it did not, and align the sigmoid output to this gold label via binary cross entropy (BCE) loss.

\subsection{Entity type predictor}

It can be easier to learn to predict the broad type of an entity to be recommended, than the specific entity itself, particularly from limited training sessions.
For each entity $e$ recommended in a `gold' training session turns $D$, we use the KB $K$ to locate triples of the form $(e, \text{is-instance-of}, t)$ --- this gives us one or more types $t$ of the recommended entity~$e$.  (In the small KBs associated with dialog data sets, entities have only one associated type.)

We reuse the encoding of conversation history $\vec{\bm{H}}$ with a different prediction head to infer the type of the entity to be recommended.  Specifically, we use a linear layer $\bm{W}\in\mathbb{R}^{D\times T}$, where $T$ is the number of types, to get logits for each type, from which we get a softmax multinomial distribution over types:
\begin{align}
\operatorname*{SoftMax}_{t\in [T]}\Big(  
 \operatorname{GPT2}(\vec{\bm{H}}) \big[ \texttt{[SUM]} \big]
\,\bm{W} 
\Big)
\end{align}
This layer is trained via standard multi-class cross-entropy loss and the type of the entity recommended in the `gold' training sessions.  During testing, we identify the top-scoring type
\begin{align}
t^*(\vec{\bm{H}}) &=
\operatorname*{argmax}_{t\in [T]} \left\{
\operatorname{GPT2}(\vec{\bm{H}}) \big[ \texttt{[SUM]} \big]
\,\bm{W}[:. t] \right\}
\end{align}
(If an entity can have more than one type, we can replace the softmax with a sigmoid that scores each type.  Adapting to more comprehensive KBs is left as future work.)

\subsection{Entity filtering and scoring}

The role of computing $t^*(\vec{\bm{H}})$ is to limit entity candidates for recommendation to those that belong to that type, viz.,
\begin{align}
E\big( t^*(\vec{\bm{H}}) \big) \subset E,
\end{align}
the full entity set in the KB.  This improves the quality of entities recommended perceptibly, even if the  final recommendation does not match ground truth.  This filtering also speeds up the scoring of entities to recommend.

The third head used to score entities wrt the unified representation uses a matrix $\bm{V}\in\mathbb{R}^{D\times E}$.
The logit score of an entity $e$ is computed as
\begin{align}
\operatorname{score}(e|\vec{\bm{H}}) = 
\operatorname{GPT2}(\vec{\bm{H}}) \big[ \texttt{[SUM]} \big] \; \bm{V}[:, e].
\end{align}
During testing, we recommend the entity
\begin{align}
\operatorname*{argmax}_{e\in E( t^*(\vec{\bm{H}}))} 
\operatorname{score}(e|\vec{\bm{H}}),
\end{align}
only if recommendation is triggered.

\paragraph{Negative sampling:}
We fine-tune all parameters of DistilBERT, GPT-2 and heads $\bm{w}, \bm{W}$ and~$\bm{V}$.
To keep training costs under control, if $E\big( t^*(\vec{\bm{H}}) \big)$ is too large, we uniformly sample a fixed number of entities from it, but always include the `gold' entity in the training conversation.
Over this subset of entities, we use the logits to define a multinomial distribution via softmax, from which we compute multi-class cross-entropy loss wrt the `gold' entity in the training conversation.

\begin{figure}
\centering\includegraphics[width=\hsize]{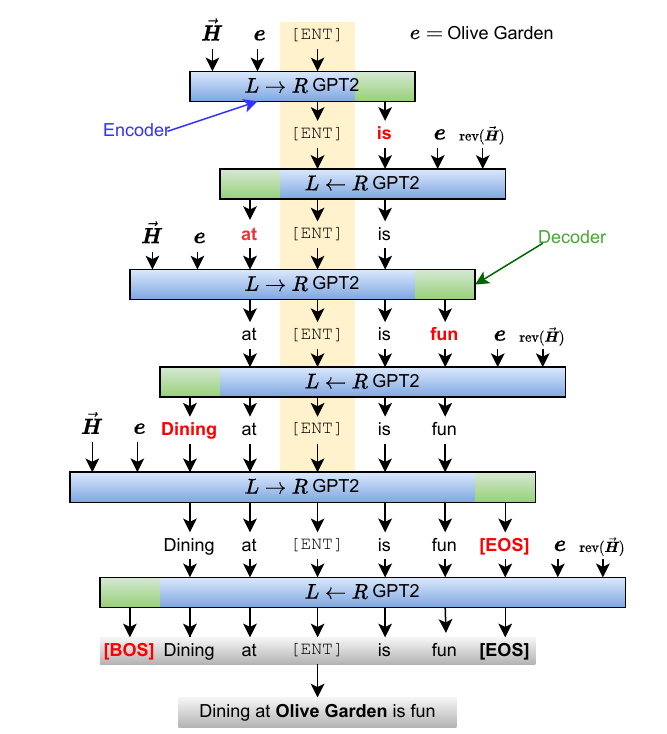}
\caption{\decodername: Bidirectional autoregressive expansion of system utterance around the entity selected for recommendation.}
\label{fig:HopSkip}
\end{figure}

\subsection{\decodername{} constrained decoder}
\label{arch:hopskip}
At this point in \sysname, the recommendation trigger has made a decision.
If the decision is to not recommend an entity, we use GPT-2 in an encoder-decoder mode, without any other constraints, to generate/decode~$\vec{a}_{M+1}$.
If the decision is to recommend, an entity $e$ is identified for recommendation. We cast the problem of generation as constrained decoding (see section \ref{sec:Relatedconst}) -- the hard constraint is that $e$ must be mentioned in the utterance~$\vec{a}_{M+1}$. 
%Constrained decoding has seen much interest recently, with general and powerful techniques such as COLD \citep{qin2022cold}, PPLM \citep{pplm} and DiffusionLM \citep{li2022diffusionlm}.
Unlike existing approaches for constrained generation which may treat constraints as soft, our constrained decoder, called \decodername, \emph{ensures} that constraint is satisfied. 
%$e^*$ is mentioned in the utterance~$\vec{a}_{M+1}$.  
Moreover, \decodername{} uses the simple structure of our constraint (must mention entity $e$ once somewhere) for a simpler solution.

\decodername{} is autoregressive, but \emph{bidirectional}.
We use GPT-2 in encoder-decoder mode, with causal attention in the decoder.
Two versions of GPT-2 are pretrained, one decoding left-to-right ($L{\rightarrow}R$) and the other 
decoding right-to-left ($L{\leftarrow}R$).
We look up the mandatory entity $e$ in the KB to find its embedding~$\bm{e}$.
Then we prime the $L{\rightarrow}R$ encoder using $\vec{\bm{H}}$ followed by~$\bm{e}$, and trigger decoding using the special token $y_0=\texttt{[ENT]}$, but generate only one token~$y_1$.
Next, we present a reversed form of history sequence $\vec{\bm{H}}$ (shown as $\operatorname{rev}(\vec{\bm{H}})$ in \ref{notation:unifiedHistory}), then $\bm{e}$, then $y_1, y_0$ to a $L{\leftarrow}R$ decoder, and trigger decoding using \texttt{[ENT]} again, emitting only one token $y_{-1}$.
We thus alternately grow the utterance one token to the right and left, centered on the entity position, as $(y_0); (y_0, y_1); (y_{-1}, y_0, y_1); (y_{-1}, y_0, y_1, y_2)$, and so on, until ``end of sentence'' \texttt{[EOS]} and ``beginning of sentence'' \texttt{[BOS]} special tokens are emitted.
During training, we use teacher forcing.

%\section{Experiments and Results}

%In this section, we discuss the datasets used, experiment setup, experiment results on both recommendation and language generation metrics, and report analysis results with ablation studies. 

\begin{table}[t]
\centering
\adjustbox{max width=\hsize}{ \tabcolsep 2pt
\begin{tabular}{crrrr}
\hline Dataset & \#Dialogs & \#Utterances & \#Entities &\#Entity Mentions \\
\hline Durecdial 2.0 & 8,241 & 127,673 & 619 & 34,708\\
ReDial & 11,348 & 182,150 & 6,924 & 51,699 \\
\hline
\end{tabular}}
\caption{Dataset Statistics}
\label{tab:datastats}
\end{table}

\section{Experimental Setup}
\label{sec:Expt}

%\subsubsection{Dataset}

\paragraph{Datasets: }We use two datasets for testing CRSs -- ReDial  \cite{redial} and a modified Durecdial 2.0 dataset \cite{liu-etal-2021-durecdial}. The ReDial dataset is developed in the context of movie recommendations, and also contains a fair amount of free-form dialog or "chit-chat." Durecdial 2.0 is originally in two languages -- English and Chinese, and on four domains -- movies, music, food, and restaurants. For our experimentation, we using the English subset of DuRecDial 2.0. We remove any intermediate annotation in this dataset (e.g., annotated goal per utterance) so that we can study end-to-end CRS design.
We curate the KB for Durecdial 2.0 from scratch, using entity linking methods as well as using entities from user profiles in the dataset. Some of the rejected annotations in the user profile are removed from our final KB. For more details of curation methodology, see Appendix \ref{appendix:kbCuration}. We will release our exact dataset along with KG for further research. The dataset statistics are in Table \ref{tab:datastats}.  

\paragraph{Evaluation metrics: }We evaluate and compare our results on two separate groups of measures, one for recommendation evaluation and other for generation quality. We used Recall@R as the metric for evaluating recommendations. It gives credit to the system when the top R ranked entities includes the ground truth entity. Specifically, we report R@1, R@10, and R@50 following previous work \citep{recallreason,kgsf,mese}. Additionally, we measure the mean reciprocal rank (MRR) of the gold entity in retrieved entity list. 
%MRR places significant emphasis on the first relevant element in the list, which is crucial in our end-to-end model as we utilize that entity for subsequent generation. 
For dialog evaluation, we utilize the BLEU score to determine the similarity between the generated response and the gold response. Furthermore, we employ Entity F1 \cite{entityf1} and multiset entity F1 \cite{multientityf1} for evaluating the entity recall in the generated utterance. Multiset F1 involves micro averaging over the multiset of entities, rather than a set -- this penalizes any model that repeats entities in an utterance (stuttering).

\paragraph{Baselines: }We compare \sysname{} with several CRSs in the literature. These include KBRD, KGSF and MESE systems (discussed in Section 2) -- they all represent top-of-the-line recent systems for the task. We also compare against the RM model, which was introduced along with the Redial dataset. RM uses an HRED-based dialog generator \cite{hred}, and a recommended module employing an auto-encoder and a sentiment analyzer. We note that for Durecdial 2.0, we do not use KGSF, as it requires two KGs (ConceptNet and entity KG), and we do not possess the annotations for ConceptNet for this dataset. 

\begin{comment}
For the evaluation of our method, we focus on two primary tasks: recommendation and response generation. Hence, our evaluation involves comparing our approach not only against existing CRS methods but also against selected baseline recommendation or conversation models that represent the area.

\textbf{RM}: Introduced in the corresponding paper along with the ReDial dataset,it comprises of a dialogue generation module based on HRED (Hierarchical Recurrent Encoder-Decoder) and a recommender module employing an auto-encoder and a sentiment analysis module.

\textbf{KBRD}: It leverages DBpedia to enrich the contextual entity semantics. It employs the Transformer architecture in the dialog generation module, utilizing KG (Knowledge Graph) information as word bias during the generation process.

\textbf{KGSF}: This model integrates both word-level (ConceptNet) and entity-level(DBPedia) knowledge graphs to improve the learning of semantic representations for user preferences. For Durecdial 2.0 dataset, we don't use this baseline as it requires two KGs to function, and we don't possess the annotations for one of the 2 KGs. 

\textbf{MESE}: The MESE model incorporates an entity metadata encoder and takes into account both item metadata and dialog context for recommendations. By considering these factors, the model learns to construct entity embeddings, enhancing its recommendation capabilities.
\end{comment}

\section{Experiments \& Results}

We seek to answer these research questions.
\begin{description}[leftmargin=.5em]
\item[Performance study:] How well does \sysname{} perform, compared to existing CRSs?
\item[Ablation:] What is the incremental value of each added module in \sysname?
\item[Error analysis:]  What are \sysname's limitations, such as incorrect recommendations, missed recommendations, or deficiencies in generating informative responses?
\end{description}

\begin{table*}[htp]
    \centering
    \resizebox{\linewidth}{!}{
    \begin{tabular}{ccccccccc}
    \toprule
    \textbf{Dataset} & \multicolumn{4}{c}{\textbf{ReDial}} & \multicolumn{4}{c}{\textbf{Durecdial 2.0}} \\
    \midrule
    Models & Recall@1 & Recall@10 & Recall@50 & MRR & Recall@1 & Recall@10 & Recall@50 & MRR \\
    \midrule
    RM & 2.42 & 14.21 & 31.57 & 0.016 & 0.040 & 2.78 & 20.81 & 0.042 \\
    KGSF & 3.39 & 17.79 & 36.52 & 0.039 & - & - & - & - \\
    KBRD & 2.87 & 16.21 & 33.81 & 0.037 & 22.21 & 48.12 & 60.12 & 0.182 \\
    MESE  & 5.54 & 25.49 & 45.64 & 0.063 & 30.30 & 65.84 & 85.39 & 0.331 \\
    \midrule
    \sysname & \textbf{5.64 }& \textbf{26.25} & \textbf{48.25} & \textbf{0.072} & \textbf{34.24} & \textbf{68.22} & \textbf{89.43} & \textbf{0.364} \\
    \bottomrule
    
    \end{tabular}}
    \caption{Automatic evaluation results on the recommendation retrieval task.}
    \label{tab:recmetrics}
\end{table*}

\begin{table*}[htp]
    \centering
    \resizebox{\linewidth}{!}{
    \begin{tabular}{ccccccccc}
    \toprule
    \textbf{Dataset} & \multicolumn{4}{c}{\textbf{ReDial}} & \multicolumn{4}{c}{\textbf{Durecdial 2.0}} \\
    \midrule
    Models & BLEU-1 & BLEU-2 & Entity F1 & Multset F1 & BLEU-1 & BLEU-2 & Entity F1 & Multiset F1 \\
    \midrule
    RM & 0.226 & 0.178 & 2.85 & 2.63 & 0.102 & 0.062 & 6.21 & 4.79 \\
    KGSF & 0.276 & 0.163 & 4.24 & 4.14 & - & - & - & - \\
    KBRD & 0.298 & 0.182 & 4.83 & 4.41 & 0.198 & 0.143 & 14.12 & 12.64 \\
    MESE  & 0.382 & 0.226 & 6.42 & 6.12 & 0.251 & 0.182 & 22.89 & 20.65 \\
    \midrule
    \sysname & \textbf{0.386} & \textbf{0.248} & \textbf{6.98 }& \textbf{6.53 }& \textbf{0.262} & \textbf{0.194} & \textbf{32.18} & \textbf{30.13} \\
    \bottomrule
    
    \end{tabular}}
    \caption{Automatic evaluation results on the response generation task.}
    \label{tab:langmetrics}
\end{table*}

% \begin{figure*}
%     \centering
%     \resizebox{\linewidth}{!}{
%     \includegraphics[width=0.3\linewidth]{plots/PlotRedial.png}
%     \includegraphics[width=0.3\linewidth]{plots/PlotDurecdial.png} 
%     \includegraphics[width=0.36\linewidth]{plots/mrr.png}
%     }
%     \caption{Automatic Evaluation Plots for Recall@X for (a) ReDial (b) Durecdial 2.0 and (c) MRR Comparisons }
%     \label{fig:recommendation}
% \end{figure*}

\subsection{Performance study}

Tables \ref{tab:recmetrics} and \ref{tab:langmetrics} presents the results of retrieval and generation metrics, respectively, for all models. We first verify that MESE is indeed the best of the baseline models. It obtains major improvements compared to other models obtaining upto 8 pt Recall@1 improvements (Durecdial) and similar BLEU-1 gains (ReDial), compared to the next best baseline. This validates our decision of building on top of MESE. Our model \sysname{} outperforms MESE with consistent gains in all datasets and metrics. It achieves about 2.5 pt Recall@50 improvement in ReDial which we attribute to joint learning with decoder, and 3-4 gains in all retrieval metrics in Durecdial 2.0, which we attribute to its type-aware entity filtering. On generation metrics, it obtains small improvements in BLEU (i.e., fluency is not hurt), but nearly 10 pt F1 gains, highlighting the importance of  constrained generation for this task. Qualitatively, we find that \sysname{} has increased informativeness, reduced chitchat and more focused on-topic responses.

\begin{table}[h]
\centering
\adjustbox{max width=\hsize}{ \tabcolsep 2pt
\begin{tabular}{lcccccc}
\toprule
     \textbf{Dataset} & \multicolumn{3}{c}{\textbf{ReDial}} & \multicolumn{3}{c}{\textbf{Durecdial 2.0}} \\
     \midrule
     Component & P & R & F1 & P & R & F1 \\
     \midrule
     Recommend.~Trigger & 77.26 & 89.47 & 82.53 & 75.22 & 92.38 & 82.52 \\ 
     Type Classifier & - & - & - & 87.24 & 91.34 & 89.25 \\
\bottomrule
\end{tabular}}
\caption{Macro-averaged scores of individual \sysname{} retriever components.}
\label{tab:classificationReport}
\end{table}

\paragraph{Trigger performance: }We further evaluate the performance of two intermediate components within \sysname. Table \ref{tab:classificationReport} reports the accuracy of recommendation trigger and type classifier. We find that the trigger has a decently high performance of about 82 F1 in both datasets, with recall being higher than precision. Since \sysname{} ensures that if trigger is 1, there exists an entity in final response, this directly increases that the informativeness of the system. Instead of generating a simple chit-chat responses, our model prefers inclusion of information from the KB, which leads to more contextually helpful responses.

%\paragraph{Entity Trigger Performance:}  From Table\,\ref{tab:classificationReport}, it can be observed that the trigger module has a higher recall as compared to its precision. This is a direct result of our focus to ensure that there exists an entity in the final generated response. As a result of this, our system has a tendency to overestimate the presence of positive instances, leading to a higher number of false positive predictions. Thus instead of generating a simple chit-chat statement, our model focuses on inclusion of information from the KB to use in our final response, which lead to more contextually expressive responses.

\begin{figure}
    \centering
    \includegraphics[width=.8\hsize]{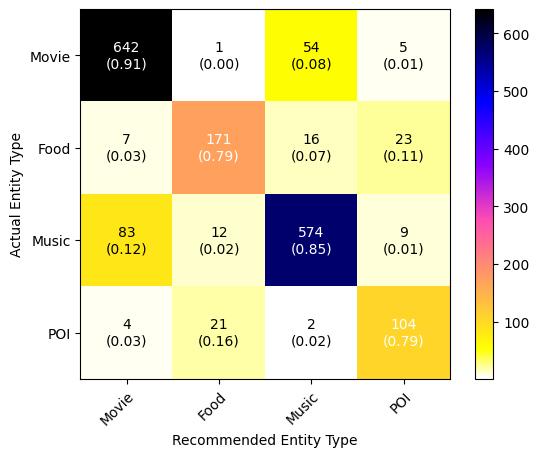}
    \caption{Confusion matrix of Type Classifier for Durecdial~2.0.}
    \label{fig:confusionMatrix}
\end{figure}

\paragraph{Type classifier performance: }Table \ref{tab:classificationReport} also shows that \sysname{} has a nearly 90\% F1 in type prediction scores, across all Durecdial 2.0 classes, \{Movies, Music, Food, POI\}. In comparison, a vanialla majority class classifier will be 40\% accurate, so overall, the classifier is doing well on the subtask. Recall that ReDial only has one entity type, hence type classifier is not relevant there. We further analyze the errors made in type predictions. We hypothesize that the model gets more confused when some types have overlapping attributes in KB. To validate this, we calculate the (pairwise) intersection of attributes between entities within frequently confused class pairs, \{food, POI\} and \{movie, music\} (see Figure \ref{fig:confusionMatrix}). We find that about 27\% of attributes were common among 75\% of the falsely classified entity pairs within these classes. 

%\paragraph{Type Classifier Performance:} 
%Our results from Table\,\ref{tab:classificationReport} demonstrate the superiority of the type classifier in accurately identifying relevant entity classes. Across the test dataset, our model achieved scores close to 90\% for precision, recall and F1 score across all classes (Movies, Music, Food, POI). We hypothesize that the occurrence of false positives and false negatives can be attributed to the overlapping nature of related attributes obtained from the knowledge base (KB). To validate this hypothesis, we conducted an analysis by calculating the intersection of attributes between pairwise entities within the frequently confused class pairs, namely \{food, POI\} and \{movie, music\}. The results were quite expected, as we found that approximately 27\% of attributes intersected among 75\% of the falsely classified entity pairs within these class pairs. Additionally, we observed an intersection of attributes of around 9\% for all the falsely detected pairs.

% \begin{figure}[h]
% \begin{minipage}[c]{0.4\linewidth}
%     \includegraphics[width=\linewidth]{plots/redial_trigger.png}
% \end{minipage}\hfill
% \begin{minipage}[c]{0.4\linewidth}
%     \includegraphics[width=\linewidth]{plots/durecdial_trigger.png}
% \end{minipage}]
% \caption{Heatmaps for the Entity Trigger Performance on (i) ReDial and (ii) Durecdial 2.0 dataset}
% \end{figure}

\paragraph{Human Evaluation:}
We perform a human evaluation to assess generated responses for: (1) Fluency, which pertains to the coherence of the responses within the context of the dialogue and the KB, and (2) Informativeness, degree of information contained in the responses compared to the gold standard response. We sample 50 dialog utterances from each of the datasets, and two judges evaluate the responses using a Likert scale \cite{likert1932technique} from 0 to 4 (both inclusive). 
Table \ref{tab:humanEval} reports the results. \sysname{} outperforms other models in all metrics, suggesting that humans indeed find \sysname{} to be an overall improvement over MESE and other models.
%The summarized results of this evaluation can be found in Table\,\ref{tab:humanEval}.

\begin{table}[ht]
\centering
\adjustbox{max width=\hsize}{  \tabcolsep 2pt
\begin{tabular}{ccccc}
\toprule
\textbf{Dataset} & \multicolumn{2}{c}{\textbf{ReDial}} & \multicolumn{2}{c}{\textbf{Durecdial 2.0}} \\
\midrule
Models & Fluency & Informativeness & Fluency & Informativeness  \\
\midrule
RM & 2.04 & 1.54 & 2.34 & 1.68 \\
KBRD & 2.98 & 2.12 & 3.16 & 2.50  \\
MESE  & 3.22 & 3.28 & 3.44 & 3.02  \\
\midrule
\sysname & \textbf{3.44} & \textbf{3.48} & \textbf{3.46} & \textbf{3.24} \\
\bottomrule
\end{tabular}}
\caption{Human evaluation results on the response-generation task.}
\label{tab:humanEval}
\end{table}

%\sysname{} performs the best in both metrics. By effectively using forward and backward decoding mechanisms using \decodername{} and leveraging information from the KB $K$, our model is able to generate more informative and expressive words and presents longer sentences in general as compared to other baselines. Also, the recommendation trigger helps identifying correct point of recommendation and identifies chit-chat statement with much ease and the decoder meanwhile maintains the fluency of the generated text. 

\begin{table*}[htp]
    \centering
    \resizebox{\linewidth}{!}{
    \begin{tabular}{lcccccccc}
        \toprule
        Model & R@1 & R@10 & R@50 & MRR & BLEU-1 & BLEU-2 & Entity F1 & Multiset F1 \\
        \midrule
        \sysname & 34.24 & 68.22 & 89.43 & 0.364 & 0.262 & 0.194 & 32.18 & 30.13 \\
        \sysname{} w/o RT & 31.14 & 65.15 & 82.52 & 0.322 & 0.255 & 0.187 & 21.16 & 20.14 \\
        \sysname{} w/o TC & 32.26 & 67.26 & 84.62 & 0.333 & 0.258 & 0.190 & 31.42 & 30.16 \\
        \sysname{} w/o RT and TC & 29.14 & 50.15 & 72.93 & 0.263 & 0.232 & 0.172 & 19.41 & 17.23 \\
        \bottomrule
        \end{tabular}
    }
    \caption{Ablation over retriever components in \sysname{}. (RT=Recommendation Trigger; TC=Type Classifier)}
    \label{tab:ablation-seperate}
\end{table*}

\begin{table}[ht]
\centering
\adjustbox{max width=\hsize}{ \tabcolsep 2pt
\begin{tabular}{lcccc}
\toprule
Decoders & BLEU-1 & BLEU-2 & Entity F1 & Multiset F1 \\
\midrule
BART & 0.212 & 0.114 & 17.20 & 15.15 \\
GPT-2 & 0.228 & 0.132 & 18.50 & 16.42 \\
GPT-2 Beam Search & 0.246 & 0.148 & 24.11 & 23.03 \\
COLD & 0.212 & 0.13 & 28.46 & 26.68 \\
NeuroLogic A*esque & 0.238 & 0.183 & 28.42 & 26.92 \\
\midrule
\decodername & \textbf{0.262} & \textbf{0.194} & \textbf{32.18} & \textbf{30.13} \\
\bottomrule
\end{tabular}}
\caption{Performance of decoders on \sysname{}.}
\label{tab:ablation-decoder}
\end{table}

\subsection{Ablation Study}

Table \ref{tab:ablation-seperate} reports the results of the ablation study, where we remove individual components of \sysname's retriever architecture to assess their incremental contribution to overall performance. Removal of recommendation trigger causes the most significant drop in both entity F1 scores. Type classifier also has a meaningful impact on final performance. Removing both components hurts all metrics considerably. 

%As shown in Table\,\ref{tab:ablation-seperate}, we first focus on the importance of individual components of our retriever architecture. Removal of the recommendation trigger causes the most significant drop in entity F1 and multiset entity F1 scores, respectively. Incorporation of the recommendation trigger thus shows the importance of the timeliness of recommendations, which leads to improved user satisfaction and engagement. The entity pruner, on the other hand, focuses more on the correct entity retrieval end. It helps the model to reduce the search space incrementally along with the trigger to output a better and more appropriate entity at a given timestep. 

We also compare our \decodername{} with other constrained and unconstrained generation approaches in literature. For this, we keep the rest of the \sysname{} architecture the same, and only change the final decoder. Table \ref{tab:ablation-decoder} shows the results. Unconstrained generation (BART, GPT2, GPT2+Beam Search) performs much worse on entity F1 metrics, since it frequently misses outputting any entity in response. Existing constrained decoding methods, COLD \cite{qin2022cold} and Neurologic A*esque \cite{neurologica*}, perform better, but still fall short of our novel \decodername{} for our task. We believe this is because left to right constrained generation has to maintain a lookahead probability distribution,  modeling that the constraint word will get added in future, which is a challenging distribution to model. However,  \decodername{}  exploits that there is only one constraint word, and hence is always using next word distribution (in either direction) without any lookahead, making it easier to model. 

%Table\,\ref{tab:ablation-decoder} depicts the effectiveness of our decoding mechanism. Instead of the other decoding mechanisms like COLD \cite{qin2022cold} and Neurologic A*esque \cite{neurologica*}, which focus on defining constraint functions through an energy function and satisfying logical constraints in CNF respectively, our model's constraint satisfaction solution is based on the effectiveness of the recommendation trigger rather than decoder due to forcing the decoder to start decoding from the constraint placeholder token. As observed from the BLEU 1/2 scores from Table\,\ref{tab:ablation-decoder}, our model focuses more on the matching the generated probability distribution of the sequence of tokens in "gold" utterances and reducing the perplexity of the output generated sequence.

\subsection{Qualitative Error Analysis}

\sysname{} shows three common failure modes. First, sometimes, agent makes no recommendation and continues general theme of the conversation:
Example, for a user query: "What movies are popular these days?" the agent responds with: "\textit{I'm not sure, but have you watched any good movies recently?}". We believe this is due to the cold-start problem, where the model has had insufficient interaction to respond with a specific entity.

Second, agent makes an incorrect recommendation. Often, the system does give a good recommendation but one that is different from the gold recommendation. Example, it may output a different movie of the same actor. In fewer cases, it confuses type of the prediction, example, instead of outputting a movie \textit{Fly Me To Polaris} by an actor \textit{Cecilia Cheung}, it outputs a song \textit{Enjoy the Moonlight Alone}, which is sung by the actor. Another example is the entity \textit{Sichuan roasted fish} which gets replaced by some restaurant recommendation, where fish is served.

%, but it is incorrect due to KG entity overlap: From the confusion matrix as shown in Fig. \ref{fig:confusionMatrix}, it is clear that there seems to be two cases where the classifier typically confuses the prediction, which are; movies and music; food and POI(restaurant). An example for which can be seen in entity pairs like : (\textit{Enjoy the Moonlight Alone}; \textit{Fly Me To Polaris}) where the theme of the conversation is \textit{Cecilia Cheung}; who is the singer of the former entity and actor of the latter entity. Another example is the entity \textit{Sichuan Roasted Fish} which is typically replaced by some restaurant recommendations where fish is served.

Third,  agent makes a correct recommendation, but the generated sentence is not as informative as the gold. As an example for a user query about popular restaurants in city center: the agent responds with: "You should try 'Restaurant X.' It's really good.". Although, grammatically and contextually correct, the generated sentence gets a lower BLEU compared to the gold response: "A popular restaurant in the city center is 'Restaurant X.' It offers a diverse menu with both local and international cuisines, and the ambience is top-notch.", due to the evaluation dependence on unigrams and bigrams.

\section{Conclusion}
\label{sec:End}

We present \sysname, an end-to-end CRS that improves upon the quality of entities recommended and the times/turns of conversation at which they are recommended, while simultaneously improving the fluency of the system utterances (using a novel application of conditional generation).  To achieve these, it uses KB information about entities and a classifier to trigger a recommendation, filters entities by types relevant to the conversation history and a bidirectional autoregressive decoder to generate an utterance around the recommended entity. 
Experiments with two benchmarks and recent baselines show the superiority of \sysname. Further analysis shows typical failure modes. We release our code and data for future research.

\clearpage

% limitations not included in 8-page limit

\section{Limitations}
\label{sec:Limit}

Recommender systems make decisions with social consequences; as they promote some products or services in preference to others, biases must be avoided and fairness enhanced.  We leave this important aspect to future work.
Given the nature of publicly available CRS data sets, the entity catalogs are small, homogeneous, and are nowhere as vast and diverse as giant KBs like \href{https://wikipedia.org/}{Wikipedia} and \href{https://wikidata.org/}{WikiData}, leave alone product catalogs maintained by leading e-commerce providers.  Such comprehensive KBs will provide additional challenges.
Finally, we can think of many other variations on history encoding, a comprehensive comparison of these is left as future work.

% Entries for the entire Anthology, followed by custom entries
\bibliography{anthology,custom}
\bibliographystyle{acl_natbib}

\appendix

\twocolumn[
\begin{center}
\Large\bfseries\ztitle \\
(Appendix)
\end{center}
]

\section{Sample interactions}

\begin{figure*}[h]
\centering
\includegraphics[width=\textwidth]{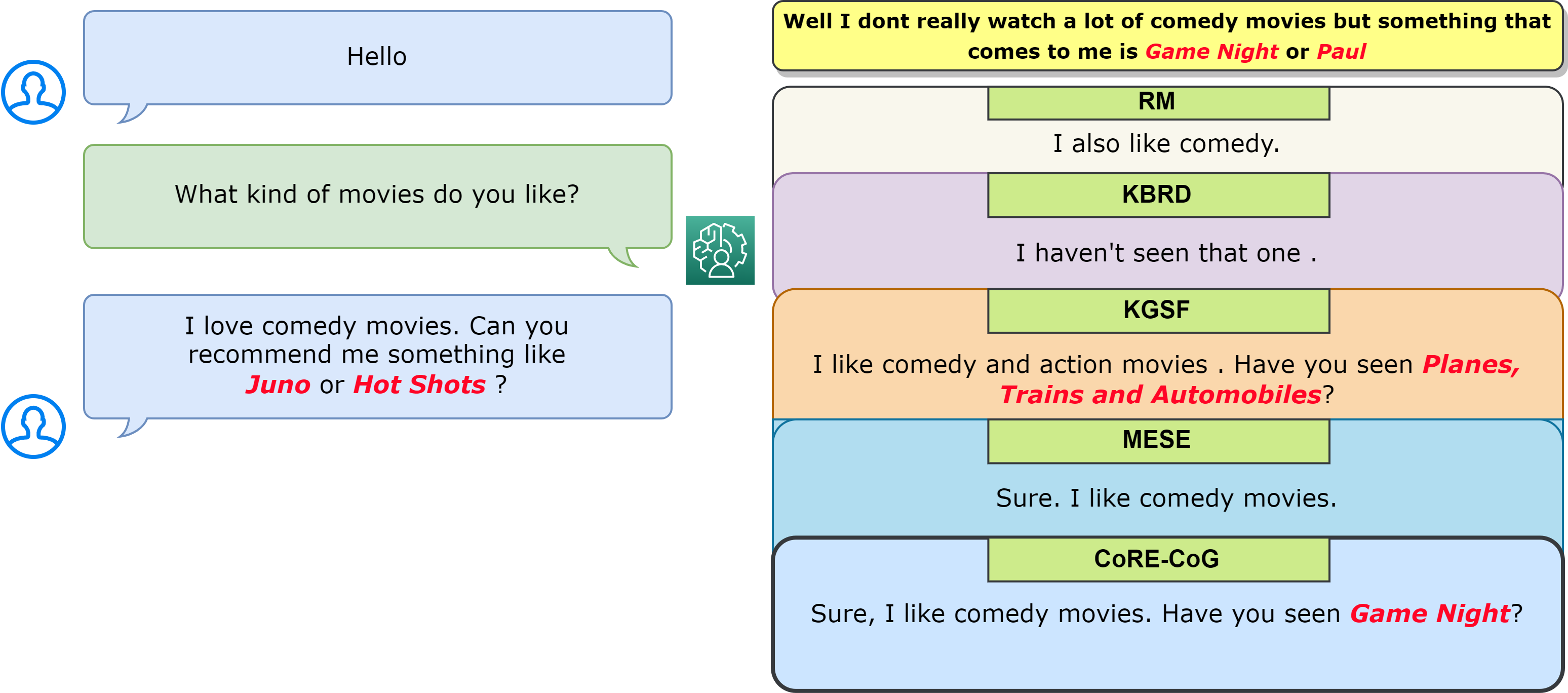}
\caption{Sample conversation response comparison for different baselines.  The yellow box shows a gold utterance.}
\label{fig:model-example}

\vspace*{\floatsep}% https://tex.stackexchange.com/q/26521/5764

\centering
\includegraphics[width=\textwidth]{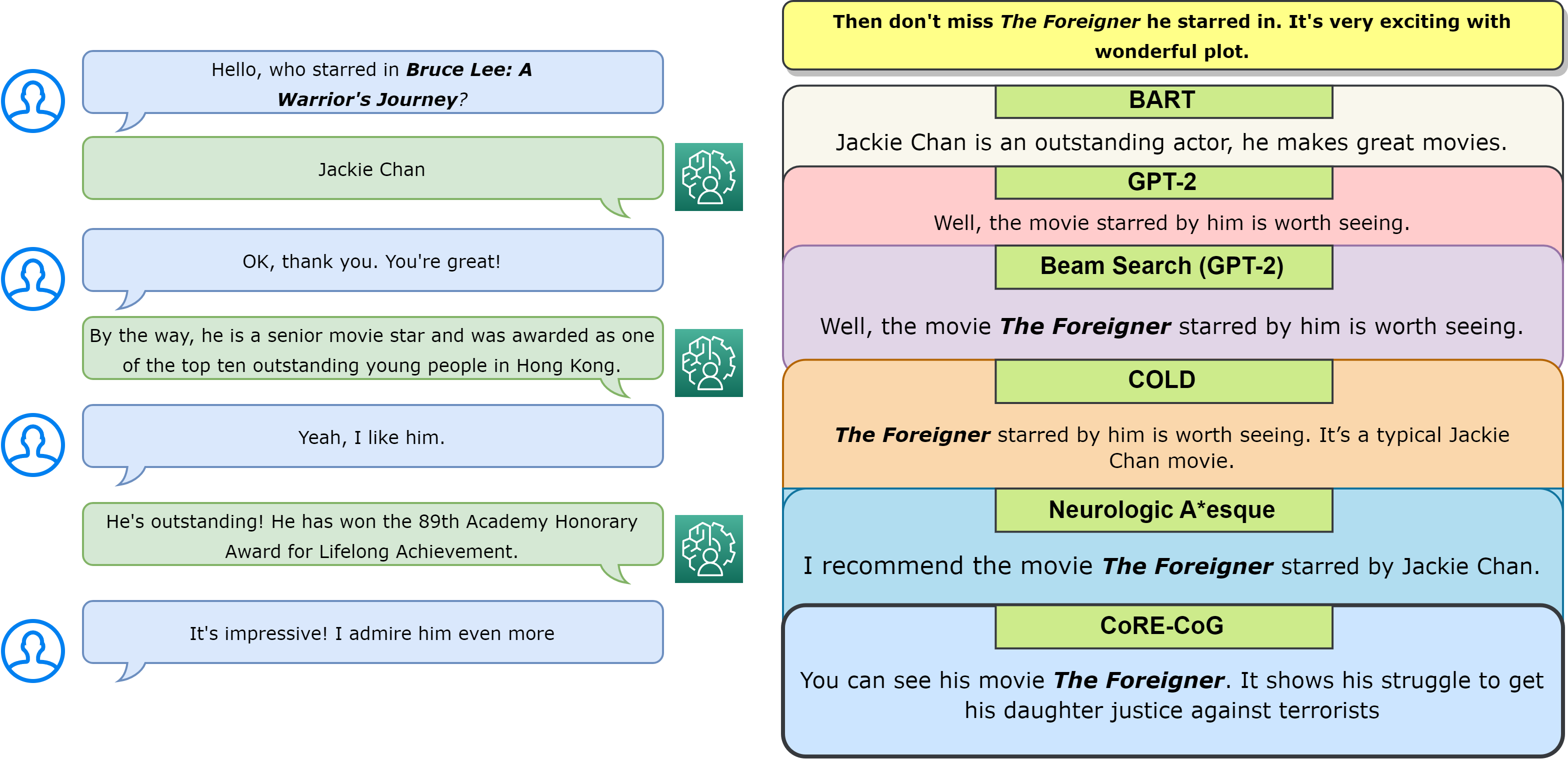}
\caption{Sample conversation when using different decoding mechanisms. The yellow box shows a gold utterance.}
\label{fig:ablation-decoder-example}
\end{figure*}

Shown below are two examples of conversations that compare \sysname{}. Figure.\ref{fig:model-example} presents an example from the ReDial dataset, while Fig.\ref{fig:ablation-decoder-example} presents an example from the Durecdial 2.0 dataset. For each Figure, the left part shows the flow of the conversation and right part shows conversational responses.

\section{KB curation for Durecdial~2.0}
\label{appendix:kbCuration}

The Knowledge Base for the Durecdial 2.0 dataset, integrated at the back-end of \sysname{}, was built from scratch. Firstly, various attributes of different entity types were identified, like genre and cast for movies, description and main ingredients for food, and so on. Let us observe this with an example. Take the example of \textit{Baked Scallion Pancakes}. We add a short description of 2-3 sentences as \textit{While traditional scallion pancakes are usually pan fried in oil, these pancakes crisped up nicely in the oven, no doubt because of the perforations in the mat, which allow air to circulate around the dough}. We also add another attribute, the main ingredients as \textit{scallions, flour, ginger, garlic, soy sauce}. These pieces of information were picked off from a variety of databases and sources. For some relevant attributes, we scraped the information from various verified databases and web-searches off the internet.\footnote{\url{https://www.themoviedb.org/}, \url{https://www.dbpedia-spotlight.org/}, \url{https://www.allmusic.com}}

\begin{algorithm}
\caption{Collect relevant attributes of entities.}
\begin{algorithmic}[1]
\Procedure{CollectAttributes}{EntitySet $E$, SourceSet $S$, WebSearchResults $W$}
    \For{each entity $e_i$ in $E$}
        \State Initialize empty set $A_i$
        \State Obtain class $c_i$ of $e_i$
        \If{$c_i$ exists in $S$}
            \State Collect attributes from source $S$
        \Else
            \State Curate attributes from multiple web search results $W$
        \EndIf
    \EndFor
\EndProcedure
\end{algorithmic}
\end{algorithm}

\section{Implementation Details}

We use GPT-2 \cite{gpt2} model as the model backbone  for dialog generation and retrieval. For the forward generation task, it contained 12 layers, 768 hidden units, 12 heads with 117M parameters. For our backward generation backbone, we used a pretrained model \cite{reflective}, with 1536 hidden units and 6 layers, 12 heads and 248M parameters.  We also used 2 item encoders  to encoder items in entity retrieval step, respectively, each has a DistilBert \cite{sanh2019distilbert} model with 6 layers, 768 hidden units, 12 heads, with 66M parameters. We used the AdamW optimizer with learning rate set to $1\mathrm{e}{-3}$. The model was trained for 15 epochs on DurecDial 2.0 dataset and 10 epochs for ReDial dataset, and the first epoch was dedicated to warm up with a linear scheduler. We set the recall size for ReDial and Durecdial 2.0 datasets as 500 and 300 respectively and performed a grid search to get the coefficient hyperparameters  for loss functions.

\end{document}